\documentclass[11pt,a4paper]{article}
\usepackage[hyperref]{emnlp2020}
\usepackage{times}
\usepackage{latexsym}

\usepackage{microtype}

\usepackage{xcolor}
\usepackage{colortbl}
\usepackage{hhline}
\usepackage{amsmath}
\usepackage{amssymb}
\usepackage{multirow}
\usepackage{booktabs}
\usepackage{graphicx}
\usepackage{enumitem}
\usepackage{rotating}
\usepackage{tikz}
\usetikzlibrary{shapes,backgrounds}
\usepackage{url}

\aclfinalcopy 

\newcommand*\myfont{\fontfamily{bch}\selectfont}
\newcommand*\socialiqa{{\myfont Social IQa}}
\newcommand*\winogrande{{\myfont WinoGrande}}
\newcommand*\copa{{\myfont COPA}}
\newcommand*\commonsenseqa{{\myfont  CommonSenseQA}}
\newcommand*\commonsenseqashort{{\myfont  CSQA}}
\newcommand*\mctaco{{\myfont MC-TACO}}
\newcommand*\piqa{{\myfont PIQA}}
\newcommand*\comet{COMeT}
\newcolumntype{e}{>{\columncolor{blue!10}}c}
\newcolumntype{d}{>{\columncolor{blue!10}}l}
\newcommand{\specialcellleft}[2][l]{%
\begin{tabular}[#1]{@{}l@{}}#2\end{tabular}
}
\newcolumntype{P}[1]{>{\raggedright\arraybackslash}p{#1}}

\title{Unsupervised Commonsense Question Answering with Self-Talk}

\author{Vered Shwartz$^{1,2}$,  Peter West$^{1,2}$, Ronan Le Bras$^1$, Chandra Bhagavatula$^1$, and Yejin Choi$^{1,2}$
       \\
       $^1$Allen Institute for Artificial Intelligence\\
       $^2$Paul G. Allen School of Computer Science \& Engineering, University of Washington\\
       {\tt \{vereds,peterw,ronanlb,chandrab,yejinc\}@allenai.org} \\
       }

\date{}

\begin{document}
\maketitle

\begin{abstract}
Natural language understanding involves reading between the lines with implicit background knowledge. Current systems either rely on pre-trained language models as the sole implicit source of world knowledge, or resort to external knowledge bases (KBs) to incorporate additional relevant knowledge. We propose an unsupervised framework based on \emph{self-talk} as a novel alternative to multiple-choice commonsense tasks. Inspired by inquiry-based discovery learning \cite{bruner1961act}, our approach inquires language models with a number of information seeking questions such as \textit{``what is the definition of ...''} to discover additional background knowledge. Empirical results demonstrate that the self-talk procedure substantially improves the performance of zero-shot language model baselines on four out of six commonsense benchmarks, and competes with models that obtain knowledge from external KBs. While our approach improves performance on several benchmarks, the self-talk induced knowledge even when leading to correct answers is not always seen as helpful by human judges, raising interesting questions about the inner-workings of pre-trained language models for commonsense reasoning. 

\end{abstract}

\section{Introduction}
\label{sec:intro}
Human level natural language understanding involves reading between the lines and relying on implicit background knowledge. Consider the sentence: \textit{Alice let Bob stand in front of her at the concert}. Using physical and social commonsense -- (i) Bob and Alice want to see the stage, and (ii) If Bob is taller, they would block Alice's view -- one can infer that Alice is taller than Bob. Such examples are ubiquitous across natural language understanding (NLU) tasks such as reading comprehension \cite{hirschman1999deep} and recognizing textual entailment \cite{dagan2013recognizing}, and even more so in tasks dedicated to commonsense reasoning such as the Winograd schema challenge \cite{levesque2012winograd}.  
Most current NLU models rely on pre-trained language models \cite[LMs; e.g.][]{gpt2,bert,t5}. The standard practice is to fine-tune a pre-trained LM in a supervised manner on task-specific data. Alternatively, LM score is used to rank answer choices in a zero-shot setup 
\cite{wang-etal-2019-make,Bosselut2019DynamicKG}. In both setups, pre-trained LMs yield improved performance upon prior methods, greatly due to the world knowledge that such LMs capture, having been trained on massive texts \cite{petroni-etal-2019-language,davison-etal-2019-commonsense}.

Despite the performance boost, LMs as knowledge providers suffer from various shortcomings: (i) \emph{insufficient coverage}: due to reporting bias, many trivial facts might not be captured by LMs because they are rarely written about \cite{gordon2013reporting}. (ii) \emph{insufficient precision}: the distributional training objective increases the probability of non-facts that are semantically similar to true facts, as in negation \cite[``birds cannot fly''; ][]{kassner2019negated}. LMs excel in predicting the semantic category of a missing word, but might predict the wrong instance in that category (e.g., depending on the phrasing, BERT sometimes predicts \emph{red} as the color of a dove). Finally, (iii) \emph{limited reasoning capabilities}: it is unclear that LMs are capable of performing multiple reasoning steps involving implicit knowledge. 

To increase the coverage of high-precision world knowledge and facilitate multi-hop reasoning by making intermediate reasoning steps explicit, prior work incorporated KBs \cite[e.g. ConceptNet;][]{speer2012representing} and knowledge-informed models into LM-based models \cite{xia2019incorporating,Bosselut2019DynamicKG,chen2019incorporating}. 
    
\begin{figure*}[t]
\centering
\frame{
    \includegraphics[width=.9\linewidth]{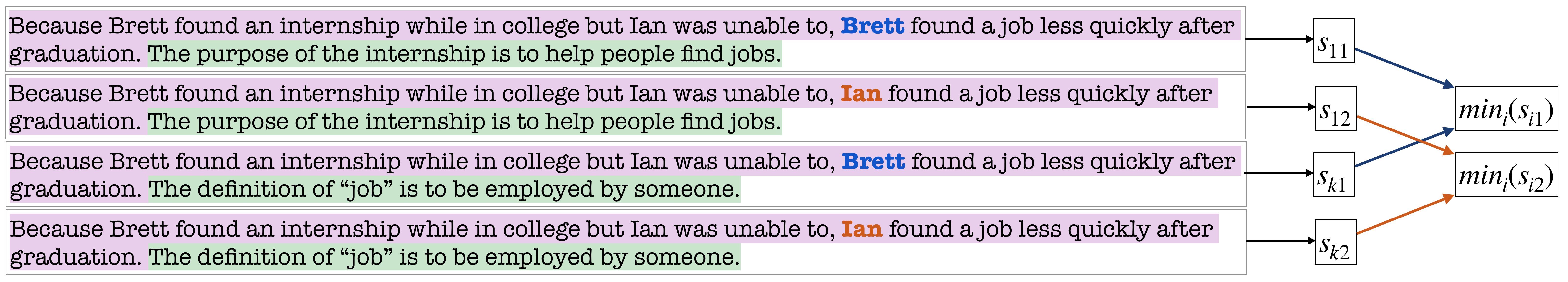}
}
\caption{Model illustration for \winogrande. Each answer choice (Brett, Ian) is assigned to the concatenation of the context and a clarification. The score for each choice is the best LM score across clarifications (2 in this case).}
\label{fig:classifier}
\end{figure*}
\vspace{-5pt}

In this paper, we study pre-trained LMs as an alternative to external KBs in providing knowledge to commonsense question answering tasks. We propose an unsupervised model that uses an LM as the answer scorer, and a (possibly different) LM as a knowledge source. We formulate the process of obtaining relevant knowledge as a \emph{self-talk}, inquiry-based discovery learning \cite{bruner1961act}, with the following steps: 1) seeking out knowledge by generating natural-language ``clarification questions'' conditioned on a given context, 2) generating their corresponding answers (``clarifications''), and 3) incorporating the clarifications as additional context.

Our model does not rely on external knowledge or additional supervision. Yet, we show that on 4 out of 6 tasks it substantially improves upon a zero-shot baseline that relies on LM score alone and performs on par, and sometimes better than, models that use external knowledge sources. 

Integrating external knowledge warrants discerning relevant and helpful facts for solving a particular instance. LMs further require identifying that a clarification is factually-correct. We show that even among the clarifications that helped the prediction, humans perceived many as unhelpful or even incorrect, demonstrating that LM-based models often solve problems correctly for seemingly incorrect reasons. Our results call for future research on robust and correct knowledge integration to LM-based question answering systems.

\section{Tasks}
\label{sec:tasks}
We focused on the multiple-choice question answering tasks detailed below. Each instance consists of an optional context, an optional question, and several answer choices. 

\paragraph{\copa: Choice of Plausible Alternatives} \cite{gordon-etal-2012-semeval}: Asking about either a plausible cause or a plausible result, among two alternatives, of a certain event expressed in a simple sentence. 

\paragraph{\commonsenseqa: commonsense Question Answering} \cite{talmor-etal-2019-commonsenseqa}: General questions about concepts from ConceptNet. To increase the challenge, the distractors are related to the target concept either by a relationship in ConceptNet or as suggested by crowdsourcing workers.

\paragraph{\mctaco: Multiple Choice Temporal commonsense} \cite{zhou2019going}: Questions about temporal aspects of events such as ordering, duration, frequency, and typical time. The distractors were selected in an adversarial way using BERT.\footnote{To make this task compatible with the other tasks, we only kept a single correct answer per instance, making our results not comparable to previously reported results.}

\paragraph{\socialiqa: Social Interaction Question Answering} \cite{sap-etal-2019-social}: Questions regarding social interactions, based on the ATOMIC dataset \cite{sap2019atomic}. Contexts describe social interactions and questions refer to one of a few aspects (e.g. the subject's motivation, following actions, etc.). The answers were crowdsourced. 

\paragraph{\piqa: Physical Interaction Question Answering} \cite{Bisk2020}: Questions regarding physical commonsense knowledge. Contexts are goals derived from an instruction website, typically involving less prototypical uses of everyday objects (e.g., using a bottle to separate eggs). The answers were crowdsourced, and an adversarial filtering algorithm was used to remove annotation artifacts.\footnote{Word associations and dataset-specific features that are not informative for the task are identified by a strong baseline and removed \cite{gururangan2018annotation,zellers-etal-2018-swag}.}

\paragraph{\winogrande} \cite{sakaguchi_winogrande:_2019}: A large-scale version of WSC that exhibits less bias thanks to adversarial filtering and use of placeholders instead of pronouns. As opposed to WSC that was curated by experts, \winogrande~was crowdsourced with a carefully designed approach that produces diverse examples which are trivial for humans. 

\section{Models}
\label{sec:approach}
\begin{figure*}[t]
\centering
\frame{
    \includegraphics[width=.9\linewidth]{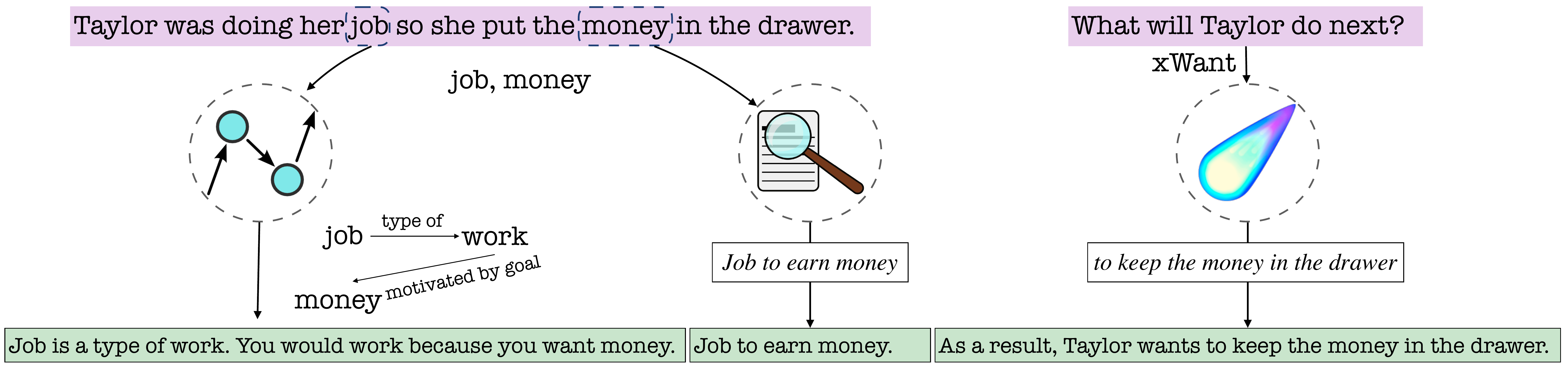}
}
\caption{Generating a single clarification using ConceptNet, Google Ngrams, and \comet{} (\socialiqa~instance).}
\label{fig:extracting_knowledge}
\end{figure*}

A given instance consists of an optional context $c$, an optional question $q$, and answer choices: $a_{i = 1}^k$. We first describe the baseline model, which makes the prediction based on the instance alone (\S\ref{sec:baseline}). We then describe a knowledge-informed model that relies on external resources (\S\ref{sec:model_from_clarifications_from_external_resources}). Finally, we discuss our self-talk model, which uses a pre-trained LMs to produce clarifications (\S\ref{sec:inquiry_based_model}). 

\subsection{LM-only Baseline}
\label{sec:baseline}

We use a pre-trained language model $\mathbb{LM}_s$ to score the plausibility of different text fragments. We experiment with the various LMs provided by the transformers package \cite{Wolf2019HuggingFacesTS}: GPT \cite{gpt}, GPT2 \cite[][all sizes]{gpt2}, a distilled GPT2 \cite{sanh2019distilbert}, and XLNet \cite[][both sizes]{xlnet}. 

We assign each of the answer choices $a_i$ into the combination of the context and the question, and obtain $\text{opt}_i = \operatorname{combine}(c, q, a_i)$. The $\operatorname{combine}$ function is computed differently for each task. For example, in \copa, where the question might be either about the cause or the effect of the context, we create the following texts for cause: ``[context]. \textit{As a result}, [choice]'' and for effect: ``[context]. \textit{The cause for it was that} [choice]''. 

We denote the score of each answer choice as $\operatorname{score}(a_i) = \operatorname{CE}(\text{opt}_i)$, where $\operatorname{CE}$ is cross-entropy loss defined as:\\ 
$\operatorname{CE}(t_1 ... t_n) = -\frac{1}{n} \sum_{i=1}^n{{\log_2 p_{\mathbb{LM}_s}(t_i \mid t_1 ... t_{i-1})}}$.\\ 
\noindent We predict the $a_i$ with the lowest score as the correct answer, which is the most likely option according to $\mathbb{LM}_s$: $y = \operatorname{argmin}_i{\operatorname{score}(a_i)}$. 

\subsection{Baseline Model with External Knowledge}
\label{sec:model_from_clarifications_from_external_resources}

In the setup illustrated in Figure~\ref{fig:classifier}, each instance consists of an additional clarification list: $CL = \{cl_1, ..., cl_m\}$. Those are text fragments containing potentially relevant knowledge for solving the instance. For example, the clarification ``\textit{The purpose of the internship is to help people find jobs}'' might help answering the question ``\textit{which of Brett and Ian found a job less quickly after graduation?}''. We don't expect all the clarifications to be relevant and helpful for answering the main question. Instead, the model relies on the single clarification that increases its belief of a certain answer choice. Thus, the score of each answer choice is selected as the score of the text containing the clarification that most supports it, i.e., whose combination with it yields the minimal loss: $\operatorname{score}(a_i) = \operatorname{min}_{cl \in CL} \operatorname{CE}(opt_i + cl)$.\\ 
\noindent Again we predict $y = \operatorname{argmin}_i{\operatorname{score}(a_i)}$.

We extract clarifications from the following sources, exemplified in Figure~\ref{fig:extracting_knowledge}.

\paragraph{ConceptNet.} Similarly to previous work, we extract relation paths between words from the context and the question, and words from the answer choices. Since we incorporate the knowledge into the model as text, we convert each ConceptNet relation to a natural language template as in \newcite{davison-etal-2019-commonsense}. We limit the path length to 2 edges in order to maintain high precision. 

\paragraph{Corpus.} For pairs of words from the context and question and from the answer choices, we extract their joint occurrences (with minimum frequency of 100) in Google N-grams \cite{brantsweb}. This yields text fragments of up to 5 words rather than well-formed sentences, with the potential of describing the relationship between the two words \cite{shwartz-dagan-2018-paraphrase}.

\paragraph{\comet{}.} \comet{} \cite{bosselut-etal-2019-comet} is a knowledge base construction model trained on the ATOMIC resource \cite{sap2019atomic} which consists of everyday situations along with multiple commonsense dimensions such as their causes, effects, pre- and post-conditions, etc. We generate all the dimensions unless we can generate specific relations that are more likely to help. Specifically, in \socialiqa, we heuristically try to understand which type of relation in \comet{} the question asks for. In \copa, we use the pre-condition relations for cause questions (\texttt{xIntent}, \texttt{xNeed}) and the post-condition relations for effect questions (\texttt{xEffect}, \texttt{xReact}, \texttt{xWant}, \texttt{oEffect}, \texttt{oReact}, \texttt{oWant}). When possible, we replace \texttt{personX} with the syntactic subject of the context or the question. 

\subsection{Self-talk Model}
\label{sec:inquiry_based_model}

\begin{figure}[t]
\centering
\frame{
    \includegraphics[width=.9\linewidth]{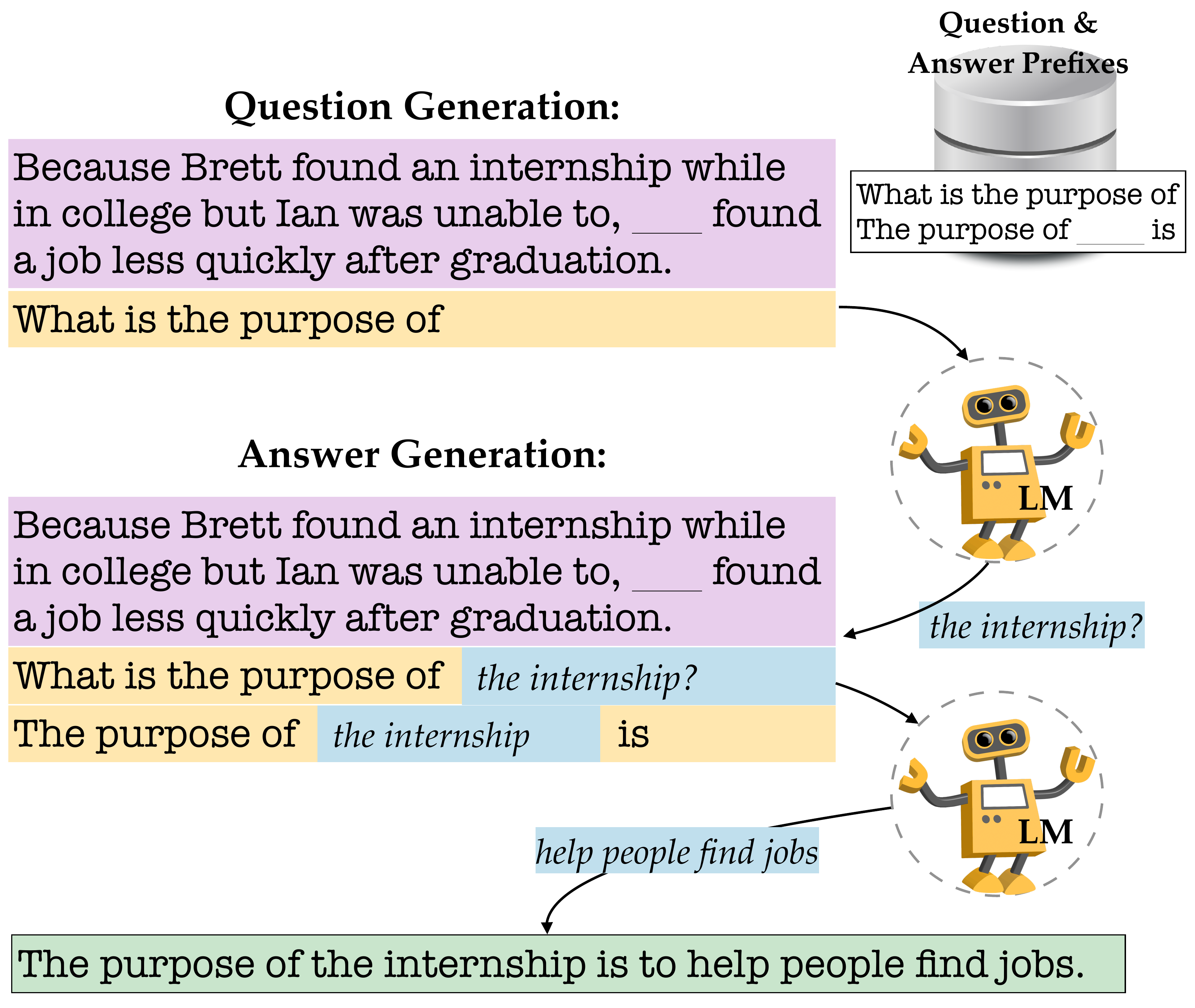}
}
\caption{Generating a clarification with LM: 1) Generate a question, conditioned on the context (pink) and question prefix (yellow). 2) Generate an answer, conditioned on the context, generated question and a corresponding answer prefix. The clarification is a concatenation of the answer prefix and generated text (green).}
\label{fig:winogrande_lm}
\end{figure}
    
Our proposed model makes the prediction identically to Figure~\ref{fig:classifier}, but extracts the clarifications from pre-trained LMs. We treat the knowledge extraction from LMs as a process of self-asking clarification questions about the context and ``discovering'' their answers. Figure~\ref{fig:winogrande_lm} exemplifies this process for \winogrande~with a generator language model $\mathbb{LM}_g$. For the sake of simplicity, the illustration depicts the process of generating a single pair of clarification question and answer.

We start by generating multiple clarification questions conditioned on the context, by 1) concatenating one of several question prefixes, which we curated for each task (e.g. ``What is the purpose of'', see Table~\ref{tab:prefixes} in the appendix); and 2) generating 5 questions for each prefix using Nucleus sampling with $p=0.2$, i.e., sampling from the top 20\% tokens \cite{holtzman2019curious}.\footnote{$p = 0.2$ is significantly lower than the standard value of $p = 0.9$ in the literature. We optimized for factual correctness, and our preliminary experiments have shown that lower $p$ values produce texts that are more faithful to the LM training corpus, at the price of being more bland.} We limit the question length to up to 6 additional tokens.

For each well-formed question that we obtained at the previous step, e.g. ``\textit{What is the purpose of the internship}?'', we generate multiple answers using a similar method. Each question prefix corresponds to an answer prefix. We use the concatenation of the context, generated clarification question, and answer prefix as the prompt for generating an answer (clarification). We limit the answer length to 10 generated tokens, and use Nucleus sampling with $p=0.5$. We generate 10 answers for each clarification question and keep all the well-formed clarifications. Note that the clarification questions themselves are only means to generate the clarifications, and they are not used by our model.\footnote{In some datasets, an instance consists of a question. In this case, we can use the instance question as a ``clarification'' question and generate additional clarification questions similar to it. For example, the \socialiqa{} context ``\textit{Austin fought for Quinn's life, but they eventually died on the operating table.}'', the LM answers the question ``\textit{Why did Austin do this?}'' directly with: ``\textit{Austin did this because they wanted to keep him alive}'' (the correct answer is ``\textit{Because Austin wanted to save Quinn}'').} 

Since we did not train the clarification generator to ask sensical, relevant, and helpful questions, nor did we train the answer generator to generate coherent and factually correct answers, we can assume that some of the generated clarifications do not provide useful information to the model. 

\section{Results}
\label{sec:results}
\begin{table*}[t]
    \scriptsize
    \centering

\setlength\tabcolsep{4pt}
\begin{tabular}{llllllllllll}
& \textbf{\comet{}} & \textbf{ConceptNet} & \textbf{Google Ngrams} & \textbf{GPT} & \textbf{Distil-GPT2} & \textbf{GPT2} & \textbf{GPT2-M} & \textbf{GPT2-L} & \textbf{GPT2-XL} & \textbf{XLNet} & \textbf{XLNet-L}\\
\textbf{\copa{}} & \cellcolor{green!51.25} 10.25 & \cellcolor{green!34.37} 6.87 & \cellcolor{green!37.50} 7.50 & \cellcolor{green!36.25} 7.25 & \cellcolor{green!26.87} 5.37 & \cellcolor{green!35.62} 7.12 & \cellcolor{green!36.87} 7.37 & \cellcolor{green!21.87} 4.37 & \cellcolor{green!38.75} 7.75 & \cellcolor{green!34.37} 6.87 & \cellcolor{green!36.87} 7.37\\
\textbf{\commonsenseqashort{}} & \cellcolor{green!1.94} 0.39 & \cellcolor{red!16.12} -3.23 & \cellcolor{red!1.50} -0.30 & \cellcolor{red!20.19} -4.04 & \cellcolor{red!18.94} -3.79 & \cellcolor{red!17.88} -3.58 & \cellcolor{red!15.44} -3.09 & \cellcolor{red!16.31} -3.26 & \cellcolor{red!18.25} -3.65 & \cellcolor{red!19.56} -3.91 & \cellcolor{red!17.75} -3.55\\
\textbf{\mctaco{}} & \cellcolor{green!9.50} 1.90 & \cellcolor{green!16.75} 3.35 & \cellcolor{green!17.62} 3.53 & \cellcolor{green!11.81} 2.36 & \cellcolor{green!12.94} 2.59 & \cellcolor{green!15.75} 3.15 & \cellcolor{green!12.81} 2.56 & \cellcolor{green!15.31} 3.06 & \cellcolor{green!14.62} 2.92 & \cellcolor{green!9.19} 1.84 & \cellcolor{green!8.75} 1.75\\
\textbf{\socialiqa{}} & \cellcolor{green!13.69} 2.74 & \cellcolor{green!6.06} 1.21 & \cellcolor{green!7.44} 1.49 & \cellcolor{green!8.56} 1.71 & \cellcolor{green!9.37} 1.87 & \cellcolor{green!8.31} 1.66 & \cellcolor{green!8.75} 1.75 & \cellcolor{green!9.75} 1.95 & \cellcolor{green!11.19} 2.24 & \cellcolor{green!8.69} 1.74 & \cellcolor{green!8.94} 1.79\\
\textbf{\piqa{}} & \cellcolor{green!18.87} 3.77 & \cellcolor{green!20.37} 4.07 & \cellcolor{green!21.81} 4.36 & \cellcolor{green!20.06} 4.01 & \cellcolor{green!18.06} 3.61 & \cellcolor{green!19.00} 3.80 & \cellcolor{green!19.44} 3.89 & \cellcolor{green!19.38} 3.88 & \cellcolor{green!19.81} 3.96 & \cellcolor{green!19.12} 3.82 & \cellcolor{green!20.50} 4.10\\
\textbf{\winogrande{}} & \cellcolor{green!0.06} 0.01 & \cellcolor{red!0.06} -0.01 & \cellcolor{red!0.56} -0.11 & \cellcolor{green!0.63} 0.13 & \cellcolor{red!0.87} -0.17 & \cellcolor{red!0.13} -0.03 & \cellcolor{red!0.19} -0.04 & \cellcolor{green!0.19} 0.04 & \cellcolor{green!0.38} 0.08 & \cellcolor{red!0.50} -0.10 & \cellcolor{red!1.25} -0.25\\
\end{tabular}

    \caption{Relative improvement upon the zero-shot baseline in terms of development accuracy, for each knowledge source averaged across LMs for each dataset.}
    \label{tab:ks_improvement}
\end{table*}

\begin{table}[!t]
    \scriptsize
    \centering
    \setlength\tabcolsep{1.5pt}
\begin{tabular}{l d d d e e e e e}
\toprule
\rowcolor{white}
\textbf{Dataset} & \textbf{Model} & \textbf{LM} & \textbf{Knowledge} & \textbf{Dev} & \textbf{Test} \\ 
\rowcolor{white}
& & & \textbf{Source} & \textbf{Acc.} & \textbf{Acc.} \\ 
\midrule
\rowcolor{white}
\multirow{6}{*}{\copa} & Majority & & & 55.0 & \\ 

 & Baseline & Distil-GPT2 & & 53.0 & \\ 
 & Ext. Knowledge & GPT2-L & \comet{} & 69.0 & \\ 
 & Self-talk & Distil-GPT2 & Distil-GPT2 & 66.0 & \\ 

\rowcolor{white}
 & Pre. Sup & T5 & & & 94.8 \\ 
 \rowcolor{white}
 & Human & & & & 100.0 \\ 
\midrule
\rowcolor{white}
 & Majority & & & 20.9 & \\ 

 & Baseline & GPT-L & & 37.2 & 34.0 \\ 
 
{\myfont Common} & Ext. Knowledge & GPT-XL & \comet{} & 39.7 & 36.2 \\ 
{\myfont SenseQA} & Self-talk & GPT-L & GPT-M & 32.4 & 26.9 \\ 

\rowcolor{white}
 & Pre. Sup & Albert ensemble & & 83.7 & 76.5 \\ 
 \rowcolor{white}
 & Human & & & 88.9 & 88.9 \\ 
\midrule
\rowcolor{white}
 & Majority & & & 40.3 & 43.0 \\ 

{\myfont MC} & Baseline & GPT2-M &   & 53.1 & 50.6 \\ 
{\myfont TACO} & External Knowledge & GPT2-XL & \comet{} & 58.8 & 55.6 \\ 
 & Self-talk & GPT2-XL & GPT2-XL & 59.9 & 58.0 \\ 

\midrule
\rowcolor{white}
 & Majority & & & 33.6 & 33.7 \\ 

 & Baseline & GPT2-L & & 41.1 & 41.1 \\ 
 & \comet-CGA$^{*}$ &  & \comet{} & 49.6 & 51.9 \\
{\myfont Social} & Ext. Knowledge & GPT2-XL & \comet{} & 47.5 & 45.3 \\ 
{\myfont IQa} & Self-talk & GPT2-XL & GPT2-L & 46.2 & 43.9 \\ 

\rowcolor{white}
 & Pre. Sup & RoBERTa-large & & 76.6 & 77.1 \\ 
 \rowcolor{white}
 & Human & & & 86.9 & 84.4 \\ 
\midrule
\rowcolor{white}
\multirow{6}{*}{\piqa} & Majority & & & 50.5 & 50.4 \\ 

 & Baseline & GPT2-XL &   & 62.6 & 63.4 \\ 
 & Ext. Knowledge & GPT2-XL & \comet{} & 69.6 & 68.4 \\ 
 & Self-talk & GPT2-XL & GPT2-M & 70.2 & 69.5 \\ 

\rowcolor{white}
 & Pre. Sup & RoBERTa-large & & 79.2 & 77.1 \\ 
 \rowcolor{white}
 & Human & & & 94.9 & 94.9 \\ 
\midrule
\rowcolor{white}
 & Majority & & & 50.4 & 50.4 \\ 

 & Baseline & GPT2-XL & & 54.8 & 54.8 \\ 
{\myfont Wino} & Ext. Knowledge & GPT2-XL & \comet{} & 55.4 & 53.7 \\ 
{\myfont Grande} & Self-talk & GPT2-XL & GPT & 54.7 & 55.1 \\ 

\rowcolor{white}
 & Pre. Sup$^{**}$ & T5 & & 86.5 & 84.6 \\ 
 \rowcolor{white}
 & Human & & & 94.1 & 94.0 \\ 
\bottomrule
\end{tabular}
    \caption{Best setup for each model type, according to development accuracy (excluding unpublished leaderboard submissions). Test accuracy is reported when labels are available or leaderboard submission was possible. $^{*}$\comet-CGA \cite{Bosselut2019DynamicKG} is a zero-shot model performing probabilistic inference over generated inferences from a \comet{} model trained on GPT2. $^{**}$ \cite{lin2020tttttackling}.}
    \label{tab:best_results}
\end{table}

Table~\ref{tab:best_results} displays the performance of the best model in each category according to the development accuracy. We report the performance of the following models: majority baseline, LM baseline (Baseline), LM-based model with external knowledge (Ext. Knowledge), Self-talk, supervised models from prior work when applicable (Pre. Sup), and human performance. Our zero-shot models are highlighted in purple. As expected, the overall performance is worse for the zero-shot models compared to the state-of-the-art supervised models, but they perform substantially better than the majority baselines on most tasks, with the exception of \winogrande~where they only slightly outperform it. Among the LM-based models, self-talk performs on par or within a few points from the external knowledge model.

\paragraph{Best Knowledge Source.} Among the knowledge informed models, \comet{} achieves the best performance across tasks. This likely happens because \comet{} can dynamically generate predictions for any context, while the other two knowledge sources are static and lack coverage. 


Table~\ref{tab:ks_improvement} shows the relative improvement in accuracy points compared to the zero-shot baseline, for each knowledge source averaged across LMs for each dataset. Interestingly, the relative improvement is fairly uniform across knowledge sources, but it varies substantially across tasks. While some tasks benefit from any added knowledge, others benefit from none.   

We also experimented with combining the clarifications from all the knowledge sources, which didn't prove beneficial except for \mctaco~(where it added +7.9 points to the dev accuracy, bringing it to 66.7). We assume that some resources added noise, making the whole smaller than the sum of its parts. 

\section{Analysis}
\label{sec:analysis}
\begin{figure*}[t]
    \centering
    \scriptsize
    \setlength\tabcolsep{4pt}
\begin{tabular}{llllllllllll}
& \textbf{COMET} & \textbf{ConceptNet} & \textbf{Distil-GPT2} & \textbf{GPT2} & \textbf{GPT2-M} & \textbf{GPT2-XL} & \textbf{GPT2-L} & \textbf{GPT} & \textbf{XLNet} & \textbf{XLNet-L}\\
\textbf{\winogrande{}} & \cellcolor{purple!36.00} 72.00 & \cellcolor{purple!21.90} 43.80 & \cellcolor{purple!18.00} 36.00 & \cellcolor{purple!30.60} 61.20 & \cellcolor{purple!41.50} 83.00 & \cellcolor{purple!34.00} 68.00 & \cellcolor{purple!35.55} 71.10 & \cellcolor{purple!33.95} 67.90 & \cellcolor{purple!36.35} 72.70 & \cellcolor{purple!41.65} 83.30\\
\textbf{\socialiqa{}} & \cellcolor{purple!45.00} 90.00 & \cellcolor{purple!28.00} 56.00 & \cellcolor{purple!33.00} 66.00 & \cellcolor{purple!37.00} 74.00 & \cellcolor{purple!36.00} 72.00 & \cellcolor{purple!38.00} 76.00 & \cellcolor{purple!38.00} 76.00 & \cellcolor{purple!40.00} 80.00 & \cellcolor{purple!18.00} 36.00 & \cellcolor{purple!26.00} 52.00\\
\textbf{\mctaco{}} & \cellcolor{purple!33.00} 66.00 & \cellcolor{purple!6.25} 12.50 & \cellcolor{purple!13.15} 26.30 & \cellcolor{purple!23.40} 46.80 & \cellcolor{purple!31.00} 62.00 & \cellcolor{purple!28.00} 56.00 & \cellcolor{purple!27.00} 54.00 & \cellcolor{purple!21.90} 43.80 & \cellcolor{purple!25.00} 50.00 & \cellcolor{purple!16.65} 33.30\\
\textbf{\piqa{}} & \cellcolor{purple!36.00} 72.00 & \cellcolor{purple!20.00} 40.00 & \cellcolor{purple!19.00} 38.00 & \cellcolor{purple!31.00} 62.00 & \cellcolor{purple!36.00} 72.00 & \cellcolor{purple!30.00} 60.00 & \cellcolor{purple!33.00} 66.00 & \cellcolor{purple!17.50} 35.00 & \cellcolor{purple!37.50} 75.00 & \cellcolor{purple!16.65} 33.30\\
\textbf{\commonsenseqashort{}} & \cellcolor{purple!33.00} 66.00 & \cellcolor{purple!27.60} 55.20 & \cellcolor{purple!22.20} 44.40 & \cellcolor{purple!24.35} 48.70 & \cellcolor{purple!33.00} 66.00 & \cellcolor{purple!36.00} 72.00 & \cellcolor{purple!32.00} 64.00 & \cellcolor{purple!50.00} 100.00 & - & \cellcolor{purple!24.05} 48.10\\
~\\
\textbf{\winogrande{}} & \cellcolor{blue!30.00} 60.00 & \cellcolor{blue!21.90} 43.80 & \cellcolor{blue!20.00} 40.00 & \cellcolor{blue!12.25} 24.50 & \cellcolor{blue!23.40} 46.80 & \cellcolor{blue!23.00} 46.00 & \cellcolor{blue!26.65} 53.30 & \cellcolor{blue!19.65} 39.30 & \cellcolor{blue!22.75} 45.50 & \cellcolor{blue!16.65} 33.30\\
\textbf{\socialiqa{}} & \cellcolor{blue!38.00} 76.00 & \cellcolor{blue!21.00} 42.00 & \cellcolor{blue!14.00} 28.00 & \cellcolor{blue!24.00} 48.00 & \cellcolor{blue!18.00} 36.00 & \cellcolor{blue!21.00} 42.00 & \cellcolor{blue!25.00} 50.00 & \cellcolor{blue!25.00} 50.00 & \cellcolor{blue!11.00} 22.00 & \cellcolor{blue!14.00} 28.00\\
\textbf{\mctaco{}} & \cellcolor{blue!30.00} 60.00 & \cellcolor{blue!6.25} 12.50 & \cellcolor{blue!21.05} 42.10 & \cellcolor{blue!23.40} 46.80 & \cellcolor{blue!24.00} 48.00 & \cellcolor{blue!30.00} 60.00 & \cellcolor{blue!27.00} 54.00 & \cellcolor{blue!14.60} 29.20 & \cellcolor{blue!20.30} 40.60 & \cellcolor{blue!16.65} 33.30\\
\textbf{\piqa{}} & \cellcolor{blue!31.00} 62.00 & \cellcolor{blue!22.00} 44.00 & \cellcolor{blue!12.00} 24.00 & \cellcolor{blue!22.00} 44.00 & \cellcolor{blue!22.00} 44.00 & \cellcolor{blue!21.00} 42.00 & \cellcolor{blue!18.00} 36.00 & \cellcolor{blue!0.00} 0.00 & \cellcolor{blue!25.00} 50.00 & \cellcolor{blue!16.65} 33.30\\
\textbf{\commonsenseqashort{}} & \cellcolor{blue!24.00} 48.00 & \cellcolor{blue!43.10} 86.20 & \cellcolor{blue!25.00} 50.00 & \cellcolor{blue!25.65} 51.30 & \cellcolor{blue!27.00} 54.00 & \cellcolor{blue!31.00} 62.00 & \cellcolor{blue!29.00} 58.00 & \cellcolor{blue!40.00} 80.00 & - & \cellcolor{blue!25.95} 51.90\\
~\\
\textbf{\winogrande{}} & \cellcolor{green!17.00} 34.00 & \cellcolor{green!6.25} 12.50 & \cellcolor{green!10.00} 20.00 & \cellcolor{green!7.15} 14.30 & \cellcolor{green!17.00} 34.00 & \cellcolor{green!12.00} 24.00 & \cellcolor{green!15.55} 31.10 & \cellcolor{green!17.85} 35.70 & \cellcolor{green!13.65} 27.30 & \cellcolor{green!16.65} 33.30\\
\textbf{\socialiqa{}} & - & \cellcolor{green!10.00} 20.00 & - & - & - & - & - & - & - & -\\
\textbf{\mctaco{}} & \cellcolor{green!10.00} 20.00 & \cellcolor{green!0.00} 0.00 & \cellcolor{green!7.90} 15.80 & \cellcolor{green!11.70} 23.40 & \cellcolor{green!15.00} 30.00 & \cellcolor{green!21.00} 42.00 & \cellcolor{green!16.00} 32.00 & \cellcolor{green!15.60} 31.20 & \cellcolor{green!9.40} 18.80 & \cellcolor{green!16.65} 33.30\\
\textbf{\piqa{}} & \cellcolor{green!14.00} 28.00 & \cellcolor{green!3.00} 6.00 & \cellcolor{green!7.00} 14.00 & \cellcolor{green!8.00} 16.00 & \cellcolor{green!15.00} 30.00 & \cellcolor{green!13.00} 26.00 & \cellcolor{green!12.00} 24.00 & \cellcolor{green!2.50} 5.00 & \cellcolor{green!12.50} 25.00 & \cellcolor{green!16.65} 33.30\\
\textbf{\commonsenseqashort{}} & \cellcolor{green!15.00} 30.00 & \cellcolor{green!17.25} 34.50 & \cellcolor{green!16.65} 33.30 & \cellcolor{green!12.80} 25.60 & \cellcolor{green!23.00} 46.00 & \cellcolor{green!25.00} 50.00 & \cellcolor{green!21.00} 42.00 & \cellcolor{green!40.00} 80.00 & - & \cellcolor{green!18.50} 37.00\\
\end{tabular}
\setlength\tabcolsep{6pt}
    \caption{Ratio of clarifications considered as \textcolor{magenta}{\textbf{relevant}} (top),  \textcolor{blue}{\textbf{factually correct}} (middle), and \textcolor{green}{\textbf{helpful}} (bottom), among the useful and grammatical or understandable clarifications for each task and knowledge source. Answers in \socialiqa{} were evaluated for helpfulness when the clarification question was different from the main question.}
    \label{fig:heatmap}
\end{figure*}

\begin{figure}[t]
    \centering
    \includegraphics[width=.4\linewidth]{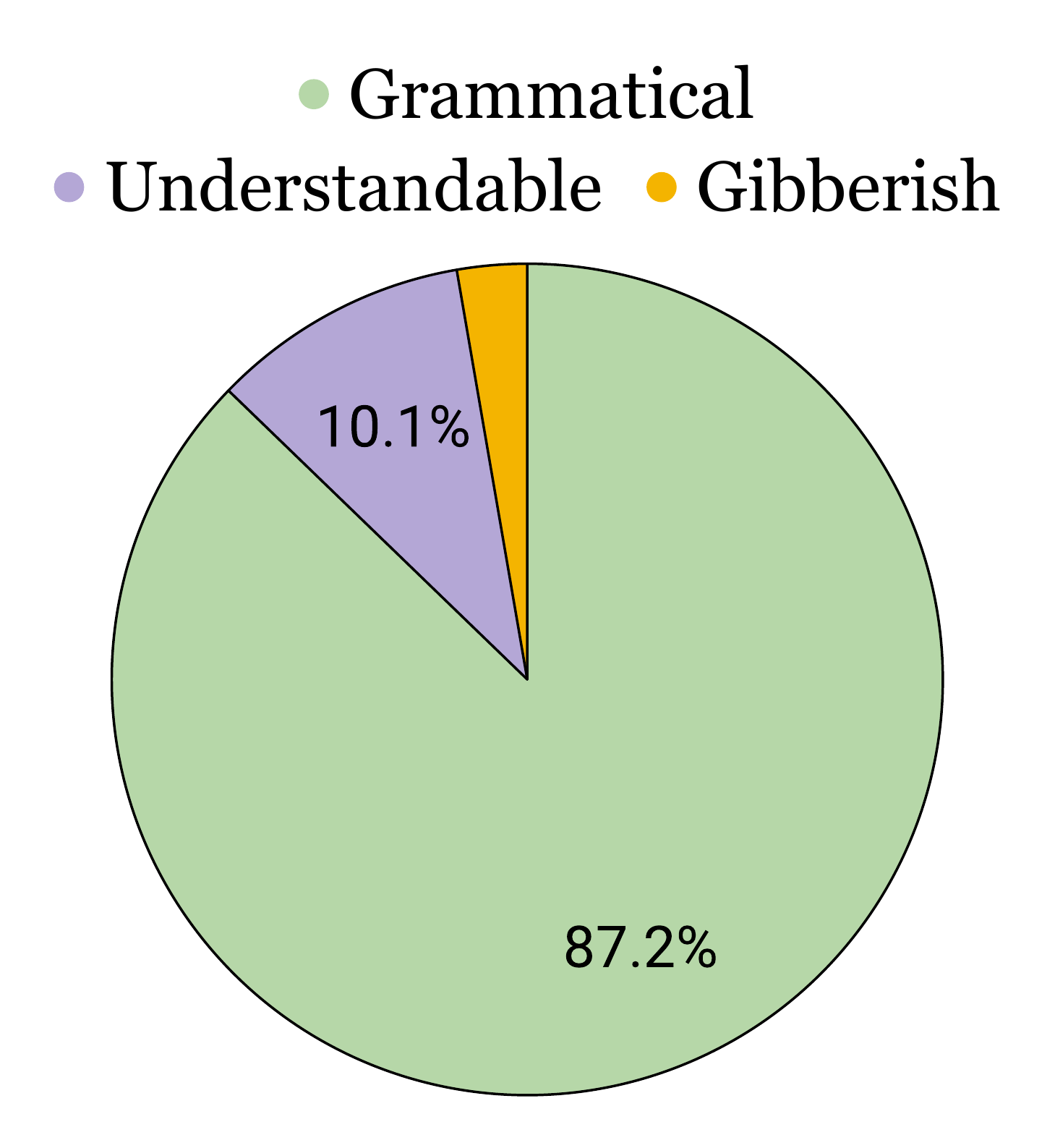}~\includegraphics[width=.6\linewidth, trim=0 2cm 0 0, clip]{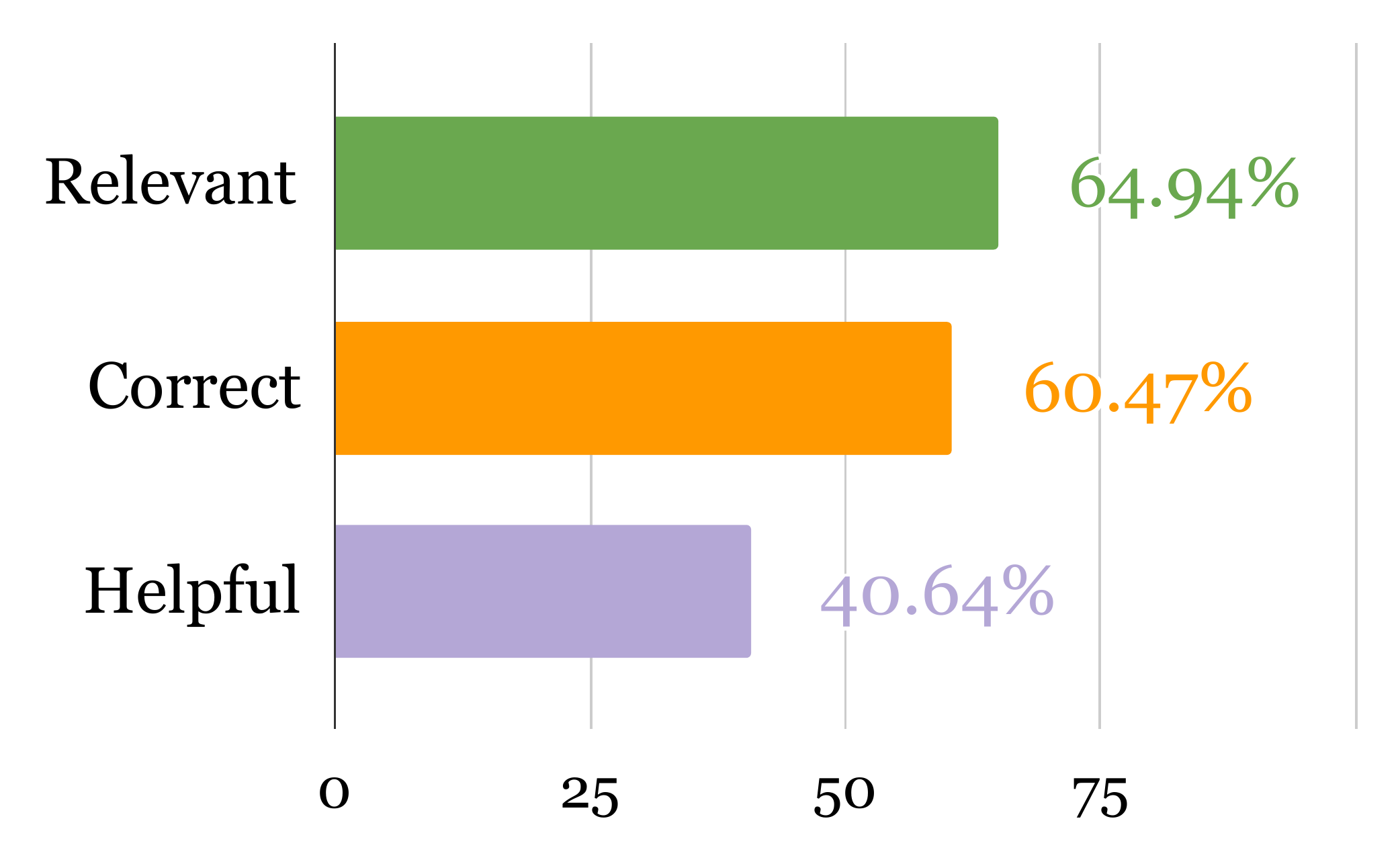}
    \caption{Human evaluation of the clarifications, aggregated across tasks and knowledge sources. \textbf{Left:} ratio of grammatical, not entirely grammatical but understandable, and completely not understandable clarifications. \textbf{Right:} percent of grammatical/understandable clarifications considered relevant, correct, and helpful.}
    \label{fig:human_eval}
\end{figure}

\begin{table*}[t]
    \scriptsize
    \centering
    \setlength\tabcolsep{1.8pt}
\begin{tabular}{>{\raggedright\arraybackslash}p{1cm} >{\raggedright\arraybackslash}p{1.2cm} >{\raggedright\arraybackslash}p{7cm} l}
\toprule
\textbf{Task} & \textbf{Source} & \textbf{Instance} &  \textbf{Clarification} \\ 
\rowcolor{lightgray} \multicolumn{4}{c}{{\textbf{Preconditions}}} \\
\multirow{5}{*}{ \commonsenseqashort{} } & \multirow{2}{*}{\parbox{1.2cm}{ ConceptNet }} & \textbf{Q:} Working on the elaborate task was taxing, it require extreme what? & \textbf{Q:} What is the relationship between `working' and `concentration'? \\
& & \textbf{Choices}: holding, \underline{concentration}, energy, job, energy & \textbf{A:} In order for working to happen, concentration needs to happen. \\
\hhline{~---}
& \multirow{3}{*}{\parbox{1.2cm}{ \comet{} }} & \textbf{Q:} When you travel you should what in case of unexpected costs? & \textbf{Q:} What do they want as a result? \\
& & \textbf{Choices}: go somewhere, energy, spend frivolously, fly in airplane, \underline{have money} & \textbf{A:} As a result, they want to make sure they have enough money.\\

\midrule
& \multirow{3}{*}{\parbox{1.2cm}{ GPT2-XL }} & \textbf{C:} [...] Mark Bailey didn't dream of becoming a judge. & \textbf{Q:} What happened first? \\
\myfont{MC-} & & \textbf{Q:} How many years did it take for Mark to become a judge? & \textbf{A:} The first thing that happened was he went to law school \\
\myfont{TACO} & & \textbf{Choices}: 63 years, [...], 7 months, 180, 7 weeks, \underline{7 years}, [...], 7 hours & and became an attorney.\\ 

\midrule


\myfont{Social} & \multirow{3}{*}{\parbox{1.2cm}{ GPT2-M }} & \textbf{C:} Carson was at a friends house and decided to play video games. & \textbf{Q:} What does Carson need to do before this?\\
\myfont{IQa} & & \textbf{Q:} What does Carson need to do before this? & \textbf{A:} Before doing that, Carson first had to get his mother's permission\\
& & \textbf{Choices:} \underline{know about their friends house}, of though about playing video games, of thought about their friend & before playing the game.\\

\rowcolor{lightgray} \multicolumn{4}{c}{{\textbf{Social Commonsense}}} \\
\myfont{Wino} & \multirow{3}{*}{\parbox{1.2cm}{ \comet{} }} & \textbf{C:} Neil looked all over the neighborhood for their missing dog until & \textbf{Q:} What does Neil feel as a result?\\
\myfont{Grande}  & & Samuel called and \underline{~~~~} got their dog back. & \textbf{A:} As a result, Neil feels happy to have the dog back.\\
& & \textbf{Choices}: \underline{Neil}, Samuel  \\

\rowcolor{lightgray} \multicolumn{4}{c}{{\textbf{Definitions and Lexical Knowledge}}} \\
\multirow{3}{*}{ \piqa{} } & \multirow{3}{*}{\parbox{1.2cm}{ \comet{} }} & \textbf{Q:} how to sanitize a toothbrush & \textbf{Q:} What does you want as a result?\\
& & \textbf{Choices}: [...] antibacterial mouthwash into a small toilet bowl [...] soak your toothbrush [...] \underline{[...] antibacterial mouthwash into a small} & \textbf{A:} As a result, you wants to get rid of germs. \\
& & \underline{cup [...] soak your toothbrush [...]} \\
\midrule
\parbox{6.5cm}{\myfont{Wino}\\ \myfont{Grande}} & GPT2-L & \parbox{6.5cm}{\textbf{C:} During the summer, I like visiting the zoo more than the aquarium because the \underline{~~~~} is inside.\\ \textbf{Choices}: zoo, \underline{aquarium}} & \parbox{6.5cm}{\textbf{Q:} What is the definition of 'zoo'? \\ \textbf{A:} The definition of a 'zoo' is an area where people can see animals in captivity.} \\
\midrule 
\parbox{6.5cm}{\myfont{Social}\\ \myfont{IQa}} & ConceptNet & 
\parbox{6.5cm}{\textbf{C:} Jan had been on vacation in Hawaii for a week. Last night, she finally returned home. \textbf{Q:} What will Jan want to do next? \textbf{Choices}: return to her family, \underline{Return to work}, Leave for another vacation.} & \parbox{6.5cm}{\textbf{Q:} What is the relationship between `vacation' and `work'?\\ \textbf{A:} Vacation is the opposite of work.} \\

\rowcolor{lightgray} \multicolumn{4}{c}{{\textbf{Functions}}} \\
\commonsenseqashort{} & DistilGPT2 & \parbox{6.5cm}{\textbf{Q:} What do professors primarily do? \\ \textbf{Choices}: master physics, state facts, wear wrinkled tweed jackets, \\ school students, \underline{teach courses}} & \parbox{6.5cm}{\textbf{Q:} What is the main function of a professor's teaching career,? \\ \textbf{A:} The main function of a professor's teaching career, is to teach students how they can improve their knowledge.} \\
\bottomrule
\end{tabular}
\setlength\tabcolsep{6pt}

    \caption{Example of relevant useful clarifications and the type of knowledge they provide.}
    \label{tab:clarification_examples}
\end{table*}

While the performance on the end task serves as an extrinsic evaluation for the quality of the generated clarifications, we are also interested in evaluating it intrinsically. From preliminary experiments we know that there is a high ratio of noisy clarifications. We thus focus on and analyze two types of clarifications: useful (\S\ref{sec:useful_eval}) and harmful (\S\ref{sec:harmful_eval}).\footnote{We omitted \copa~from the analysis due to its small size.} 

\subsection{Useful Clarifications}
\label{sec:useful_eval}

We define a clarification as \emph{useful} if (a) it is the clarification with the best LM score in its instance (i.e., the clarification used in practice); and (b) the instance was incorrectly predicted by the zero-shot baseline but correctly predicted by the self-talk model. We sampled up to 50 useful clarifications for each combination of task and knowledge source, using the best performing LM (See Table~\ref{tab:clarification_examples} in the appendix for examples). We showed crowdsourcing workers an instance along with a clarification question and its answer, and asked them: 1) whether the question is grammatical, not entirely grammatical but understandable, or completely not understandable; and if the answer was anything but ``completely not understandable'', 2) whether the question is relevant, i.e. on topic with the instance. We asked the same questions about the answer, in addition to: 3) whether the answer is factually correct or likely true; and 4) whether the answer adds helpful information to solve the instance. 

The annotation task was carried out in Amazon Mechanical Turk. To ensure the quality of annotations, we required that the workers be located in the US, UK, or Canada, and have a 99\% approval rate for at least 5,000 prior tasks. We aggregated annotation from 3 workers using majority vote. The annotations yielded moderate levels of agreement, with Fleiss’ Kappa $\kappa$ = 0.43 \cite{landis1977measurement}. Among the different categories of annotations we measured pairwise accuracy, which ranged from 60.41\% (the answer is factually correct) to 92.26\% (the question is completely not understandable).

For the sake of brevity, we focus on the analysis of the answers to the clarification questions. The left part of Figure~\ref{fig:human_eval} shows that across tasks and resources, most clarifications are grammatical or at least understandable. Among the clarifications considered grammatical or understandable, the right part of the figure shows the percentage of clarifications considered relevant, correct, and helpful. Most clarifications were considered relevant to the context and factually correct, but only 40\% on average were considered helpful. Considering that these are all clarifications that indeed helped the model, this is an interesting though not completely unexpected finding: the model utilizes knowledge that humans wouldn't consider as helpful.\footnote{Seemingly unhelpful clarifications may yet increase the LM score by adding relevant lexical cues. A manual examination of a sample of answers judged as relevant but unhelpful revealed that 53.33\% were answers for unhelpful questions, 20\% were correct but unhelpful, 16.67\% were factually incorrect, 10\% were helpful to some extent (containing knowledge deemed too trivial by the annotators), and 10\% had corresponding unanswerable instances.}

Breaking down by knowledge source, Figure~\ref{fig:heatmap} shows the ratio of clarifications considered by humans as \textcolor{magenta}{\textbf{relevant}} (top), \textcolor{blue}{\textbf{factually correct}} (middle), and \textcolor{green}{\textbf{helpful}} (bottom), for each task and knowledge source. XLNet performs worse on all measures. ConceptNet's clarifications are often judged as irrelevant likely because they are limited to a very specific type of clarification (the relationship between a pair of terms). It's not too surprising that clarifications generated by LMs were sometimes judged as factually incorrect. We also note that \comet{} generated factually correct clarifications for \socialiqa{} (which is based on ATOMIC, on which \comet{} was trained), and ConceptNet generated factually correct clarifications for \commonsenseqa{} (which is based on ConceptNet). 

Table~\ref{tab:clarification_examples} demonstrates the types of knowledge in useful and relevant clarifications, showing that pre-trained LMs do particularly well in definitions.

\begin{figure}[t]
    \centering
    \includegraphics[width=.75\linewidth]{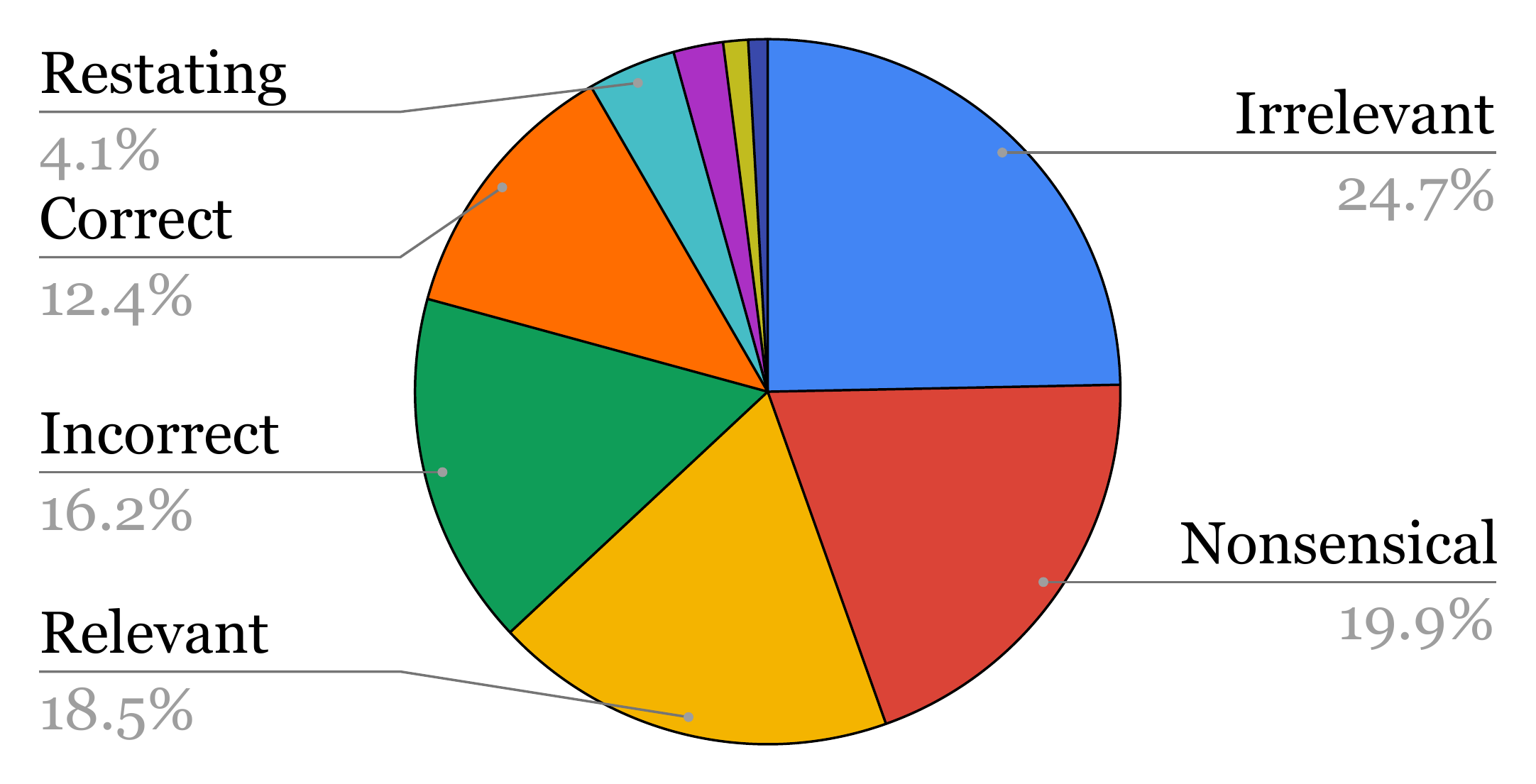}    \caption{Types of errors caused by the harmful clarifications across all tasks and knowledge sources.}
    \label{fig:error_types}
\end{figure}

\begin{table*}[t]
    \scriptsize
    \centering
    \setlength\tabcolsep{1.8pt}
\begin{tabular}{>{\raggedright\arraybackslash}p{1.5cm} >{\raggedright\arraybackslash}p{1.5cm} >{\raggedright\arraybackslash}p{7cm} >{\raggedright\arraybackslash}p{5.5cm}}
\toprule
\textbf{Task} & \textbf{Know. Source} & \textbf{Instance} &  \textbf{Clarification} \\ 
\rowcolor{lightgray} \multicolumn{4}{c}{{\textbf{Irrelevant}}} \\
\multirow{2}{*}{ \piqa } & \multirow{2}{*}{\parbox{1.5cm}{ GPT2-XL }} & \textbf{Q:} how do you sit a baby in a restaurant? & \textbf{Q:} What is the definition of ``a good time''? \\
& & \textbf{Choices}: \underline{place them in a booster seat.}, place them on the table. & \textbf{A:} The definition of ``a good time'' is not the same as what constitutes an acceptable meal. \\
\rowcolor{lightgray} \multicolumn{4}{c}{{\textbf{Nonsensical or ungrammatical}}} \\
\multirow{2}{*}{ \socialiqa } & \multirow{2}{*}{\parbox{1.5cm}{ XLNet }} & \textbf{C:} Cameron went out of their way to help a friend who was in need of help. \textbf{Q:} How would Cameron feel after helping? & \textbf{Q:} How would Cameron feel after helping? \\
& & \textbf{Choices}: Cameron then frustrated., \underline{Cameron then happy.}, Cameron then annoyed. & \textbf{A:} Cameron felt they were doing, but then he realized that. \\
\rowcolor{lightgray} \multicolumn{4}{c}{{\textbf{Relevant}}} \\
\multirow{2}{*}{ \winogrande } & \multirow{2}{*}{\parbox{1.5cm}{ GPT2-XL }} & 
The children were not vaccinated, which was fine with Betty but annoyed Mary. \underline{~~~~~} believed they made kids autistic. & \multirow{2}{*}{\parbox{5.5cm}{\textbf{Q:} What does it mean to be ``autistic''?" \\ \textbf{A:} Be ``autistic'' means to have problems in social interaction and communication skills.}} \\
& & \textbf{Choices}: \underline{Betty}, Mary \\
\rowcolor{lightgray} \multicolumn{4}{c}{{\textbf{Factually Incorrect}}} \\
\multirow{2}{*}{ \winogrande } & \multirow{2}{*}{\parbox{1.5cm}{ \comet }} & 
Derrick really liked working in Human Resources and William wanted to work at the same company, and \underline{~~~~~} subsequently offered a position. & \textbf{Q:} What will Derrick want as a result? \\
& & \textbf{Choices}: \underline{Derrick}, William & \textbf{A:} As a result, Derrick wants to go to work with them \\
\rowcolor{lightgray} \multicolumn{4}{c}{{\textbf{Correct}}} \\
\multirow{2}{*}{ \commonsenseqashort } & \multirow{2}{*}{\parbox{1.5cm}{Google Ngrams}} & \textbf{Q:} What do people usually feel when falling in love? & \textbf{Q:} - \\ 
& & \textbf{Choices}: getting married, pain, \underline{happiness}, getting married, suffering & \textbf{A:} Suffering from unrequited love.\\
\rowcolor{lightgray} \multicolumn{4}{c}{{\textbf{Restating the instance}}} \\
\multirow{2}{*}{ \commonsenseqashort } & \multirow{2}{*}{\parbox{1.5cm}{ \comet }} & \textbf{Q:} Billy set aside a block of time for having fun after work. Why might he do this? & \textbf{Q:} What will Billy want as a result? \\ 
& & \textbf{Choices}: happiness, \underline{stress relief}, pleasure, ocean, may laugh & \textbf{A:} As a result, they want to do something fun.\\
\rowcolor{lightgray} \multicolumn{4}{c}{{\textbf{Wrong Sense}}} \\

\multirow{2}{*}{ \mctaco } & \multirow{2}{*}{\parbox{1.5cm}{ConceptNet}} &
\textbf{C:} [...] Islam thrived as a strong, male-dominated religion of individuality [...] preaching brotherhood [...]. & \multirow{2}{*}{\parbox{5.5cm}{\textbf{Q:} What is the relationship between brotherhood and alcohol? \\ \textbf{A:} You are likely to find brotherhood in a fraternity house. You are likely to find alcohol in a fraternity house.}} \\ 
& & \textbf{Q:} What happened after Islam became popular in the region? \\
& & \textbf{Choices}: they drank liquor, it died off, \underline{it expanded even further}, ~~~~~ they drank alcohol, it died out, it died down \\

\bottomrule
\end{tabular}
\setlength\tabcolsep{6pt}

    \caption{An example for each of the error types among the harmful clarifications.}
    \label{tab:error_examples}
\end{table*}

\subsection{Harmful Clarifications}
\label{sec:harmful_eval}

Symmetrically, we also study the \emph{harmful} clarifications. A clarification is harmful if (a) it is the clarification with the best LM score in its instance; and (b) the instance was correctly predicted by the zero-shot baseline but incorrectly predicted by the self-talk model. We sampled up to 25 harmful clarifications from the predictions of the best setup (LM and knowledge source) for each task, and manually categorized the errors into the following types. 

\begin{enumerate}[leftmargin=*,itemsep=-0.5pt]
    \item \textbf{Irrelevant}: the clarification was off topic. 
    \item \textbf{Nonsensical or ungrammatical}: the clarification was not a complete sentence, or had other grammar or meaning issues. 
    \item \textbf{Relevant}: the clarification contributed relevant knowledge but it wasn't enough for predicting the correct answer. 
    \item \textbf{Factually Incorrect}: the clarification made a factually incorrect statement, often in support of one of the distractors.
    \item \textbf{Correct}: the clarification yielded an alternative correct answer for the main instance.
    \item \textbf{Restating the instance}: the clarification repeated the context or the main question.
    \item \textbf{Wrong sense}: the clarification interpreted a word from the instance in the wrong sense. 
    \item \textbf{Dataset error}: the instance is incorrect or lacks information required for answering it correctly. 
\end{enumerate}

Figure~\ref{fig:error_types} shows the percent of each error type across all the tasks and knowledge sources. The majority of clarifications are irrelevant, ungrammatical or nonsensical, or relevant but not helpful for making the correct prediction. We judged a non-negligible 12.4\% of the clarifications as providing alternative correct answers, phrased differently from the gold answer. Table~\ref{tab:error_examples} provides an instance for each error type.\footnote{See Figures~\ref{fig:error_type_by_task} and \ref{fig:error_type_by_ks} in the appendix for a breakdown of error types by task and knowledge source.}

\section{Related Work}
\label{sec:related_work}
\subsection{External Knowledge in Neural Models}
\label{sec:incorporating_knowledge}

Approaches for incorporating external knowledge into a neural model consist of several components: (1) the task addressed; (2) neural model; (3) knowledge sources; and (4) incorporation method. Most models target tasks that require commonsense knowledge, such as the story cloze test \cite{rocstories} and machine comprehension tasks \cite{kovcisky2018narrativeqa,ostermann2018semeval,clark_think_2018,talmor-etal-2019-commonsenseqa}. The neural component has recently shifted from biLSTM to transformer-based representations, specifically pre-trained LMs \cite{bert,roberta}. 

With respect to the knowledge source, the vast majority of papers rely on ConceptNet to extract relation paths between concepts and entities identified in the input \cite[][see an example in Figure~\ref{fig:extracting_knowledge}]{speer2012representing}. Additional resources include WordNet \cite{lin-etal-2017-reasoning,wang-jiang-2019-explicit}, retrieval or statistics mind from corpora \cite{lin-etal-2017-reasoning,mitra2019additional,joshi2020contextualized}, knowledge base embeddings \cite{chen2019incorporating,xiong-etal-2019-improving}, hand-crafted rules \cite{lin-etal-2017-reasoning,tandon-etal-2018-reasoning}, and tools such as sentiment analyzers \cite{chen2019incorporating} and knowledge-informed LMs \cite{Bosselut2019DynamicKG}.

The external knowledge is typically incorporated into the neural model by learning a vector representation of the symbolic knowledge (e.g. subgraphs from ConceptNet), and attending to it via attention mechanism when representing the inputs \cite{bauer-etal-2018-commonsense,paul-frank-2019-ranking,lin-etal-2019-kagnet}. Alternative approaches include using the knowledge to score answer candidates and prune implausible ones \cite{lin-etal-2017-reasoning,tandon-etal-2018-reasoning}, and training in a multi-task setup via auxiliary tasks pertaining to knowledge \cite{xia2019incorporating}. 

To the best of our knowledge, our method is the first to generate knowledge from pre-trained language models and incorporate it as external knowledge into a question answering model. Concurrently, \newcite{latcinnik2020explaining} used one language model to generate hypotheses and another language model as an answer scorer for \commonsenseqa{}.

\subsection{Extracting Knowledge from LMs}
\label{sec:extracting_knowledge}

Pre-trained LMs such as GPT2 \cite{gpt2} and BERT \cite{bert} capture various types of world knowledge. \newcite{petroni-etal-2019-language} showed that such LMs can be used in a KB completion task over ConceptNet and Wikidata \cite{vrandevcic2014wikidata} by converting KB relations into natural language templates and querying the LM for the missing part in the triplet (concept$_1$, relation, concept$_2$). For instance, querying BERT for suitable substitutes to the mask in ``Dante was born in [MASK]'' assigns the highest probability to Rome. \newcite{davison-etal-2019-commonsense} similarly showed that BERT assigns higher scores to natural language fragments of true rather than fictitious ConceptNet triplets, and semi-automated the template creation by using GPT2 to score hand-crafted templates. 

While both works have shown somewhat promising results, other work showed that knowledge extracted from LMs is expectantly not always accurate. Specifically, \newcite{kassner2019negated} showed that negated facts are also considered likely by the LM, while \newcite{logan_baracks_2019} pointed out that LMs may over-generalize and produce incorrect facts such as ``Barack Obama's wife is Hillary''.

\subsection{Generating Questions and Explanations}
\label{sec:asking_question}

There are numerous research directions investigating automatic question generation \cite{vanderwende_importance_2008}. Motivations vary from data augmentation to QA tasks \cite{du-etal-2017-learning,dhingra-etal-2018-simple,du-cardie-2018-harvesting,sachan2018self,fabbri-etal-2020-template} through conversational machine reading \cite{saeidi-etal-2018-interpretation,pan-etal-2019-reinforced}, simplifying questions to make them more easily answerable \cite{buck2018ask,talmor-berant-2018-web,perezunsupervised}, to using questions as means for other purposes such as sentence representation and summarization \cite{guo-etal-2018-soft,potash2019playing}.  

In particular, our work is pertinent to previous work in producing clarification questions and explanations. \newcite{rao-daume-iii-2019-answer} worked on questions from forums (e.g. Stack Exchange). They proposed a model that generates clarification questions and corresponding answers for a given question, using the question's comments (clarification questions and answers) as supervision. Question-answer pairs were scored based on how much relevant information they add to the context. 

\newcite{shen2019learning} developed an active learning framework for image captioning that learns to detect uncertainty about generated words and ask natural language questions to reduce its uncertainty. A visual question answering (VQA) model provides an answer which is then used to change the caption. The framework is trained with reinforcement learning, but the gold standard captions are used during a warmup steps and the VQA model is supervised. 

\newcite{klein2019learning} proposed a joint question generation and question answering framework. They fine-tuned GPT2 on a question answering dataset to generate a question and an answer span for a given passage, and trained BERT to answer the generated question given the passage. Finally, \newcite{rajani_explain_2019} proposed a model for \commonsenseqa~ that generates explanations for its predictions. They collected human explanations and used them to fine-tune LMs to automatically generate explanations. These explanations were then added as additional inputs. The shortcoming of this approach is that it requires collecting specific human explanations for each new dataset.



\section{Discussion and Conclusion}
\label{sec:conclusion}
We presented an unsupervised framework for multiple choice commonsense tasks that generates and integrates background knowledge from pre-trained LMs. On most tasks, it performs substantially better than the baseline and similarly to a model that had access to external knowledge resources. 

We have listed several shortcomings of using pre-trained LMs as knowledge providers: (i) \emph{insufficient coverage}, (ii) \emph{insufficient precision}, and (iii) \emph{limited reasoning capabilities}. Despite their insufficient precision compared to a KB like ConceptNet, we showed that clarifications generated by LMs resulted in similar or superior empirical gains. Among the clarifications used in practice by the answer scorer, about 60\% of those that yielded a correct prediction and 12\% of those that yielded an incorrect prediction were judged by humans as factually correct. 
 
By design, our model makes a single additional reasoning step explicit, aiming to facilitate reasoning about implicit inferences. A preliminary experiment in which we incorporated clarification pairs to facilitate two hops got mixed results. An interesting future direction is to generate each clarification in response to the previous ones, in a dialogue setup \cite{saeidi-etal-2018-interpretation}. Another challenge is the ``needle in a haystack'' problem of the clarifications, and one way to address it is to develop a model that is capable of ``introspection'', specifically knowing what it doesn't know. A more structured knowledge generation might also make the combination of various knowledge sources more successful.

Filling in knowledge gaps and making implicit intermediate reasoning steps explicit is imperative going forward. We hope that our framework will facilitate future research in this area. Our code and data will be made available upon publication. Our code and data is available at \href{https://github.com/vered1986/self_talk}{github.com/vered1986/self\_talk}.


\section*{Acknowledgements}
This research was supported in part by NSF (IIS-1524371, IIS-1714566), DARPA under the CwC program through the ARO (W911NF-15-1-0543), and DARPA under the MCS program through NIWC Pacific (N66001-19-2-4031).

\bibliography{references}
\bibliographystyle{acl_natbib}

\appendix

\begin{table*}[t!]
    \scriptsize
    \centering

\setlength\tabcolsep{4pt}
\begin{tabular}{lllllllll}
& \textbf{GPT} & \textbf{Distil-GPT2} & \textbf{GPT2} & \textbf{GPT2-M} & \textbf{GPT2-L} & \textbf{GPT2-XL} & \textbf{XLNet} & \textbf{XLNet-L}\\
\textbf{\copa{}} & \cellcolor{blue!29.32} 58.64 & \cellcolor{blue!31.86} 63.73 & \cellcolor{blue!29.86} 59.73 & \cellcolor{blue!30.91} 61.82 & \cellcolor{blue!30.32} 60.64 & \cellcolor{blue!28.95} 57.91 & \cellcolor{blue!25.95} 51.91 & \cellcolor{blue!24.73} 49.45\\
\textbf{\commonsenseqashort{}} & \cellcolor{blue!13.79} 27.57 & \cellcolor{blue!12.72} 25.45 & \cellcolor{blue!12.82} 25.64 & \cellcolor{blue!13.87} 27.74 & \cellcolor{blue!15.88} 31.75 & \cellcolor{blue!15.61} 31.22 & \cellcolor{blue!10.74} 21.47 & \cellcolor{blue!10.40} 20.79\\
\textbf{\mctaco{}} & \cellcolor{blue!23.86} 47.72 & \cellcolor{blue!24.38} 48.75 & \cellcolor{blue!25.03} 50.06 & \cellcolor{blue!26.50} 52.99 & \cellcolor{blue!28.30} 56.61 & \cellcolor{blue!29.03} 58.05 & \cellcolor{blue!17.09} 34.18 & \cellcolor{blue!18.51} 37.03\\
\textbf{\socialiqa{}} & \cellcolor{blue!20.81} 41.62 & \cellcolor{blue!20.20} 40.39 & \cellcolor{blue!20.90} 41.80 & \cellcolor{blue!21.70} 43.39 & \cellcolor{blue!22.20} 44.39 & \cellcolor{blue!22.75} 45.50 & \cellcolor{blue!16.56} 33.12 & \cellcolor{blue!16.82} 33.65\\
\textbf{\piqa{}} & \cellcolor{blue!28.95} 57.91 & \cellcolor{blue!29.81} 59.63 & \cellcolor{blue!30.98} 61.95 & \cellcolor{blue!32.79} 65.57 & \cellcolor{blue!33.95} 67.89 & \cellcolor{blue!34.80} 69.59 & \cellcolor{blue!24.62} 49.24 & \cellcolor{blue!24.40} 48.80\\
\textbf{\winogrande{}} & \cellcolor{blue!26.09} 52.18 & \cellcolor{blue!25.47} 50.94 & \cellcolor{blue!25.58} 51.16 & \cellcolor{blue!25.09} 50.18 & \cellcolor{blue!26.43} 52.85 & \cellcolor{blue!27.02} 54.04 & \cellcolor{blue!24.54} 49.07 & \cellcolor{blue!24.37} 48.74\\
\end{tabular}
    \vspace{-9pt}
    \caption{Average self-talk accuracy for each LM answer scorer, averaged across knowledge sources.}
    \label{tab:lm_rankings}
\end{table*}

\section{Question and Answer Prefixes}
\label{sec:appendix_model}

We came up with question and answer prefixes by experimenting with a few generic prefixes and observing what generally yields accurate answers. For example, we observed that LMs are not very good at causal and temporal relationships but are pretty good at definitions. For the datasets whose instances include questions (e.g. \socialiqa) we also used the corresponding question prefixes. 

Table~\ref{tab:prefixes} presents the question and answer prefixes used for each task. ``\_'' in the answer prefix is replaced with the generated question (excluding the question mark), e.g. ``What is the definition of \underline{a cat?}'' yields the answer prefix: ``The definition of a cat is''. The \socialiqa{} templates correspond to \comet{} dimensions. X is replaced with the syntactic subject of the sentence.

\begin{table}[!ht]
    \tiny
    \setlength\tabcolsep{1.8pt}
\begin{tabular}{l l l}
\toprule
\textbf{Dataset} & \textbf{Question Prefix} & \textbf{Answer Prefix} \\
\midrule
\multirow{7}{*}{\specialcellleft{\copa \\ \& \\ \commonsenseqashort{} }} & What is the definition of & The definition of \_ is \\ 
& What is the main purpose of & The purpose of \_ is to \\ 
& What is the main function of a & The main function of a \_ is \\ 
& What are the properties of a & The properties of a \_ are that \\ 
& What is a & \_ is \\ 
& What happened as a result of & As a result of \_, \\ 
& What might have caused & The cause of \_ was \\ 
\midrule
\multirow{5}{*}{\specialcellleft{\myfont MC \\ TACO }} & How long did this take? & This lasted for \\
& How often does this happen? & Every \\ 
& How many times did this happen? & This happened \\ 
& What happened first? & The first thing that happened was \\ 
& What happened last? & The last thing that happened was \\
\midrule
\multirow{19}{*}{\specialcellleft{\myfont Social \\ IQa }} & What will X want to do next? & X wanted \\
& What will X want to do after? & X wanted \\
& How would X feel afterwards? & X felt \\
& How would X feel as a result? & X felt \\
& How would X feel after? & X felt \\
& How would you describe X? & X is a \\
& What kind of person is X? & X is a \\
& How would you describe X as a person? & X is a \\
& Why did X do that? & X did this because they wanted \\
& Why did X do this? & X did this because they wanted \\
& Why did X want to do this? & X did this because they wanted \\
& What does X need to do beforehand? & Before doing that, X first had to \\
& What does X need to do before? & Before doing that, X first had to \\
& What does X need to do before this? & Before doing that, X first had to \\
& What did X need to do before this? & Before doing that, X first had to \\
& What will happen to X? & X \\
& What will happen to X next? & X \\
& What will X do next? & X \\
& What did X do? & What X did was \\ 
\midrule
\multirow{9}{*}{\piqa} & How to & The way to do \_ is \\
& How do you & The way you do \_ is \\
& How can one & One can \_ by \\
& What can be used for & \_ can be used for \\
& What can one do in order to & In order to \_, one can \\
& What should you use for & For \_, you should you use \\
& What is the definition of & The definition of \_ is \\
& What are the properties of a & The properties of a \_ are that \\
& What is a & \_ is \\ 
\midrule
\multirow{6}{*}{\specialcellleft{\myfont Wino \\ Grande }} & What is the definition of & The definition of \_ is \\
& What is the main purpose of & The purpose of \_ is to \\
& What is the main function of a & The main function of a \_ is \\
& What are the properties of a & The properties of a \_ are that \\
& What is & \_ is \\
& What does it mean to & \_ means \\
\bottomrule
\end{tabular}

    \vspace{-9pt}
    \caption{Question \& answer prefixes used for each task.}
    \label{tab:prefixes}
\end{table}

\section{Best Language Model}
\label{sec:appendix_results}

Table~\ref{tab:lm_rankings} shows the average development accuracy of the LMs across the different knowledge sources. In general there is a preference to GPT-2, and in particular to the larger models, except for \copa~in which the distilled version works best. A possible explanation might be that the language model distillation reduces the likelihood of rare words \cite{tang2018adaptive}, which works well for the simple sentences in \copa. The XLNet models perform poorly, perhaps due to their smaller training corpus (16GB vs 40GB in GPT-2, both using web text).

\begin{figure*}[t!]
    \centering
    \scriptsize
    \setlength\tabcolsep{4pt}
\begin{tabular}{llllllllllllllllll}
& \textbf{COMET} & \textbf{ConceptNet} & \textbf{Distil-GPT2} & \textbf{GPT2} & \textbf{GPT2-M} & \textbf{GPT2-XL} & \textbf{GPT2-L} & \textbf{GPT} & \textbf{XLNet} & \textbf{XLNet-L} \\
\textbf{\winogrande{}} & \cellcolor{blue!47.00} 94.00 & \cellcolor{blue!46.85} 93.70 & \cellcolor{blue!46.00} 92.00 & \cellcolor{blue!41.80} 83.60 & \cellcolor{blue!46.85} 93.70 & \cellcolor{blue!48.00} 96.00 & \cellcolor{blue!44.45} 88.90 & \cellcolor{blue!42.85} 85.70 & \cellcolor{blue!40.90} 81.80 & \cellcolor{blue!41.65} 83.30\\
\textbf{\socialiqa{}} & \cellcolor{blue!48.00} 96.00 & \cellcolor{blue!45.00} 90.00 & \cellcolor{blue!47.00} 94.00 & \cellcolor{blue!46.00} 92.00 & \cellcolor{blue!47.00} 94.00 & \cellcolor{blue!47.00} 94.00 & \cellcolor{blue!47.00} 94.00 & \cellcolor{blue!47.00} 94.00 & \cellcolor{blue!25.00} 50.00 & \cellcolor{blue!31.00} 62.00\\
\textbf{\mctaco{}} & \cellcolor{blue!47.00} 94.00 & \cellcolor{blue!31.25} 62.50 & \cellcolor{blue!42.15} 84.30 & \cellcolor{blue!44.70} 89.40 & \cellcolor{blue!47.00} 94.00 & \cellcolor{blue!48.00} 96.00 & \cellcolor{blue!49.00} 98.00 & \cellcolor{blue!43.70} 87.40 & \cellcolor{blue!39.10} 78.20 & \cellcolor{blue!50.00} 100.00\\
\textbf{\piqa{}} & \cellcolor{blue!49.00} 98.00 & \cellcolor{blue!39.00} 78.00 & \cellcolor{blue!35.00} 70.00 & \cellcolor{blue!42.00} 84.00 & \cellcolor{blue!44.00} 88.00 & \cellcolor{blue!37.00} 74.00 & \cellcolor{blue!42.00} 84.00 & \cellcolor{blue!27.50} 55.00 & \cellcolor{blue!25.00} 50.00 & \cellcolor{blue!33.30} 66.60\\
\textbf{\commonsenseqashort{}} & \cellcolor{blue!47.00} 94.00 & \cellcolor{blue!48.25} 96.50 & \cellcolor{blue!44.45} 88.90 & \cellcolor{blue!44.85} 89.70 & \cellcolor{blue!45.00} 90.00 & \cellcolor{blue!49.00} 98.00 & \cellcolor{blue!48.00} 96.00 & \cellcolor{blue!50.00} 100.00 & - & \cellcolor{blue!40.70} 81.40\\
\end{tabular}
\setlength\tabcolsep{6pt}
    \caption{Ratio of clarifications considered by humans as \textbf{grammatical or understandable} among the useful clarifications for each task and knowledge source.}
    \label{fig:grammatical_heatmap}
\end{figure*}

\section{Analysis}
\label{sec:appendix_analysis}

\subsection{Useful Clarifications}
\label{sec:appendix_useful_eval}

Figure~\ref{fig:grammatical_heatmap} shows, for each task and knowledge source, the ratio of useful clarifications that were considered by humans as either grammatical or at least understandable. The majority of the helpful clarifications are considered as grammatical. The XLNet models are slightly worse in terms of grammaticality. For example, the clarification question ``\textit{What are the properties of a you sharpen a pencil,?}'' and the answer ``\textit{The properties of a you sharpen a pencil, are that it will not break or be dulled}'' generated for the \piqa{} instance ``sharpen a pencil'' by XLNet-base. Despite its grammar errors, the answer was still useful for a LM to determine the correct answer.  



\subsection{Harmful Clarifications}
\label{sec:appendix_harmful_eval}

\begin{figure}[h]
    \centering
    \includegraphics[width=\linewidth]{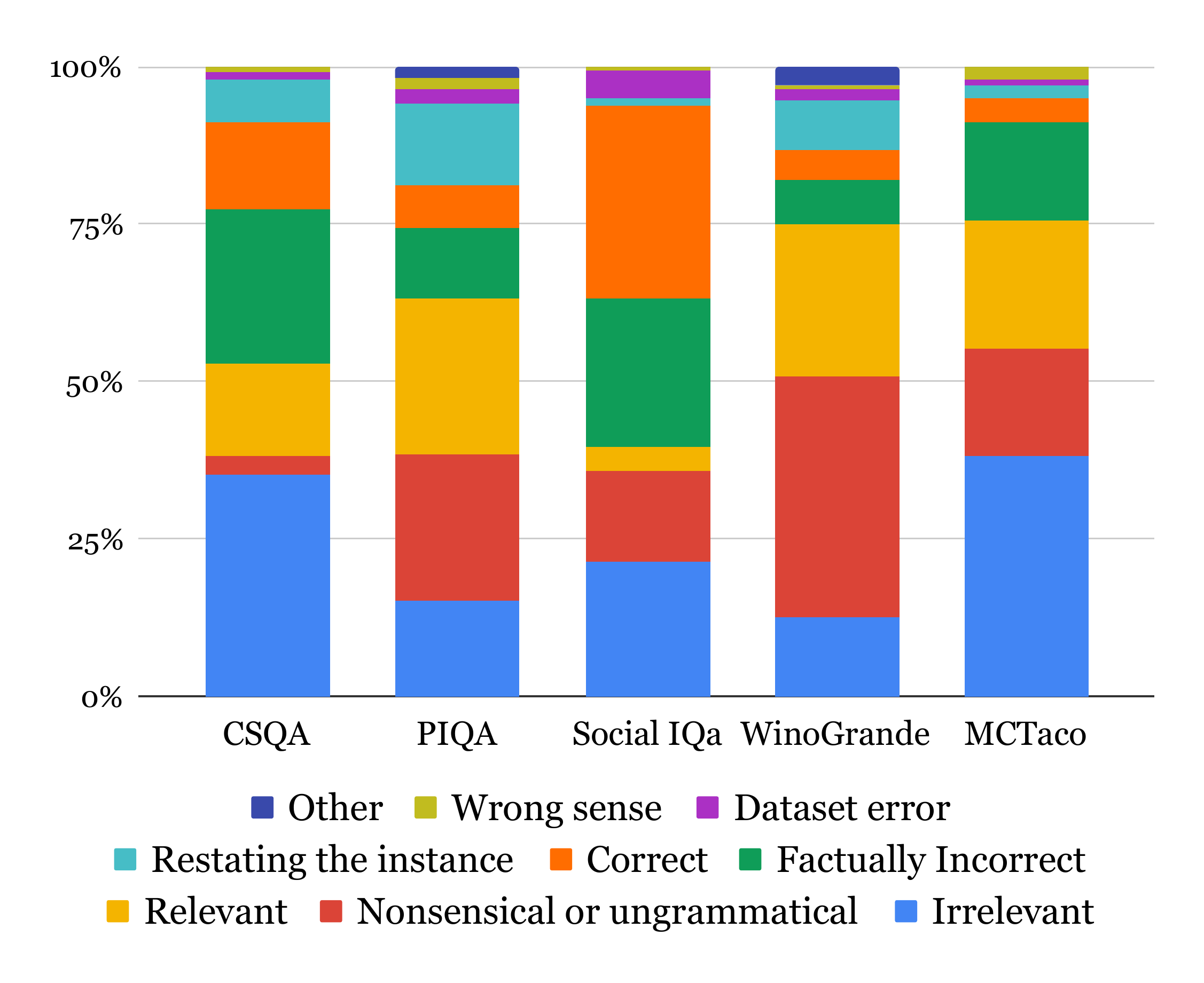}
    \caption{Types of errors caused by the harmful clarifications, for each task, across all knowledge sources.}
    \label{fig:error_type_by_task}
\end{figure}

Figure~\ref{fig:error_type_by_task} breaks down by task the type of errors found in the harmful clarifications. In \socialiqa{} and \commonsenseqa{}, many alternative correct answers are generated, but this doesn't happen in \winogrande{}, that by design only allows for one correct answer. Clarifications in \mctaco{} are more than average irrelevant. In the future, it would be interesting to investigate whether this is due to inherent lack of temporal commonsense in LMs or due to misguided attempts to extract it. 

\begin{figure}[h]
    \centering
    \includegraphics[width=\linewidth]{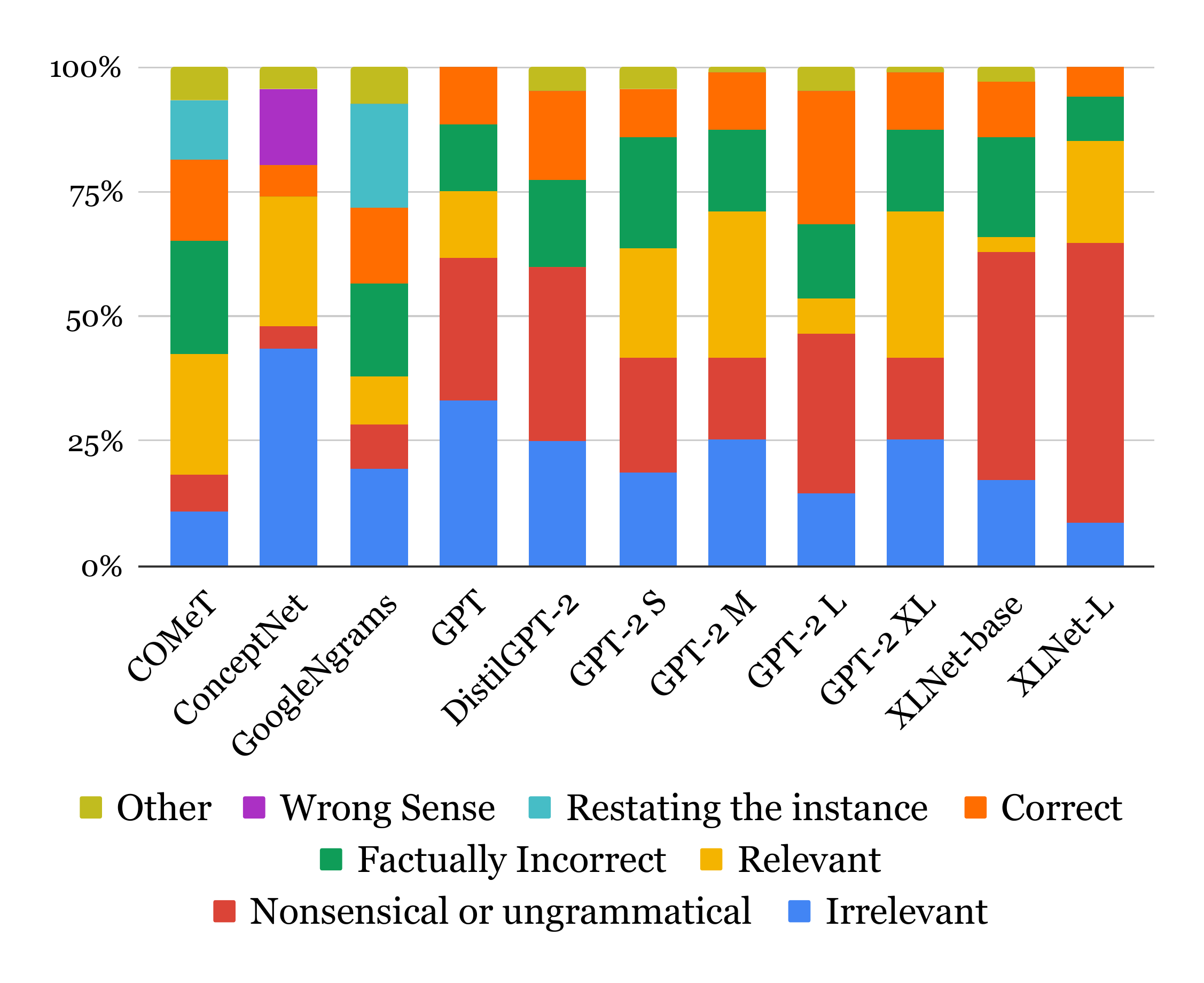}
    \caption{Types of errors caused by the harmful clarifications, for each knowledge source, across all tasks.}
    \label{fig:error_type_by_ks}
\end{figure}

Figure~\ref{fig:error_type_by_ks} similarly breaks down the errors by knowledge source. All knowledge sources except for ConceptNet make incorrect statements, but LMs also tend to make nonsensical statements, especially XLNet. ConceptNet tends to generate irrelevant clarifications (about the relationship between two unimportant terms). Being a static resource, is was also insensitive to the word senses. Google Ngrams, the only other static knowledge source, didn't suffer from this issue. This is likely because a polysemous
term $x$ related to $y$ in one of its senses wouldn't typically co-occur with $y$ in its non-related senses \cite{shwartz-dagan-2016-path}.


\end{document}